\documentclass[11pt]{article}

\usepackage[final]{acl}

\usepackage{times}
\usepackage{latexsym}

\usepackage[T1]{fontenc}

\usepackage[utf8]{inputenc}
\usepackage{multicol}

\usepackage{microtype}

\usepackage{inconsolata}

\usepackage{graphicx}

\usepackage{amsmath}
\usepackage{amsfonts}
\usepackage{fontawesome5} 

\usepackage[most]{tcolorbox}
\tcbset{
  colback=gray!5,
  colframe=gray!70,
  boxrule=0.5pt,
  arc=6.5pt,
  left=6pt,right=6pt,top=4pt,bottom=4pt,
}
\newtcolorbox{highlightbox}{}

\usepackage{xcolor}
\usepackage{colortbl}
\usepackage{tabularx}
\usepackage{enumitem}
\usepackage{bm}

\definecolor{academicblue}{RGB}{0, 76, 153}      
\definecolor{lightblue}{RGB}{198, 219, 239}      
\definecolor{academicgreen}{RGB}{0, 128, 102}    
\definecolor{lightgreen}{RGB}{210, 235, 223}     
\definecolor{academicred}{RGB}{165, 30, 55}      
\definecolor{lightred}{RGB}{244, 214, 214}       
\definecolor{academicorange}{RGB}{204, 102, 0}
\definecolor{lightorange}{RGB}{250, 232, 213}
\definecolor{darkgray}{gray}{0.3}    
\definecolor{midgray}{gray}{0.5}     
\definecolor{lightgray}{gray}{0.9}   
\definecolor{modelgray}{RGB}{240,240,240} 
\definecolor{modelbg}{gray}{0.96}
\definecolor{modelpurple}{RGB}{235,235,253}  

\definecolor{basegray}{gray}{0.45}

\definecolor{gainred}{RGB}{160, 0, 0}

\newcommand{\base}[1]{\textcolor{basegray}{#1}}

\newcommand{\our}[1]{#1}

\newcommand{\gain}[1]{\textcolor{gainred}{ (\ensuremath{\uparrow}#1)}}

\newcommand{\n}[0]{\base{--}}

\usepackage{threeparttable}
\usepackage{adjustbox}
\usepackage{booktabs}
\usepackage{multirow}

\usepackage{booktabs}
\usepackage[dvipsnames]{xcolor}
\usepackage{array}
\usepackage{xspace}
\usepackage{booktabs}
\usepackage{siunitx}
\usepackage{url}

\newcommand{\model}{\texttt{EVU}\xspace}

\title{Seeing Isn't Believing: Mitigating Belief Inertia via Active \\ Intervention in Embodied Agents}

\author{
    Hanlin Wang$^{1}$, ~~Chak Tou Leong$^{1}$, ~~Jian Wang$^{1,2 \dagger}$, ~~Wenjie Li$^{1}$ \\
    $^1$ Department of Computing, The Hong Kong Polytechnic University \\
    $^2$ College of Computer Science, Sichuan University \\
    \texttt{\{hanlin-henry.wang, chak-tou.leong\}@connect.polyu.hk} \\
    \texttt{jian51.wang@polyu.edu.hk} ~~
    \texttt{cswjli@comp.polyu.edu.hk}
}

\begin{document}
\maketitle

\renewcommand{\thefootnote}{$\dagger$}
\footnotetext[1]{Corresponding author. This work was mainly conducted at PolyU, while the author is now at Sichuan University.}
\setcounter{footnote}{0}
\renewcommand{\thefootnote}{\arabic{footnote}}

\begin{abstract}

Recent advancements in large language models (LLMs) have enabled agents to tackle complex embodied tasks through environmental interaction. 
However, these agents still make suboptimal decisions and perform ineffective actions, as they often overlook critical environmental feedback that differs from their internal beliefs. 
Through a formal probing analysis, we characterize this as \textit{belief inertia}, a phenomenon where agents stubbornly adhere to prior beliefs despite explicit observations. 
To address this, we advocate active belief intervention, moving from passive understanding to active management. 
We introduce the Estimate-Verify-Update (\model) mechanism, which empowers agents to predict expected outcomes, verify them against observations through explicit reasoning, and actively update prior beliefs based on the verification evidence. 
\model is designed as a unified intervention mechanism that generates textual belief states explicitly, and can be integrated into both prompting-based and training-based agent reasoning methods. 
Extensive experiments across three embodied benchmarks demonstrate that \model consistently yields substantial gains in task success rates. 
Further analyses validate that our approach effectively mitigates belief inertia, advancing the development of more robust embodied agents. 
Our code is available at \url{https://github.com/WangHanLinHenry/EVU}.

\end{abstract}

\section{Introduction}
\label{sec:intro}

Large language models (LLMs) have revolutionized embodied AI, enabling agents to solve increasingly complex, long-horizon tasks~\citep{huang2022language,wang2023voyager,li2024embodied}.
Effective task-solving requires not only sophisticated reasoning, but also continuous interaction with the embodied environment.
To this end, prior work has explored a variety of techniques, including inference-time iteration~\citep{yao2022react,shinn2023reflexion}, imitation learning~\citep{chen2023fireact}, and reinforcement learning~\citep{wu2025reinforced}.
The key to these methods is a tight feedback loop, in which the agent perceives the environment, interprets observations to internal states, reasons, and executes actions accordingly.
Since an agent's entire decision-making process critically depends on environmental feedback, integrating the observed information into its reasoning is crucial for task success~\citep{wang2024e2cl,fung2025embodied}.

\begin{figure}[t!]
\centering
    \includegraphics[width=0.99\linewidth]{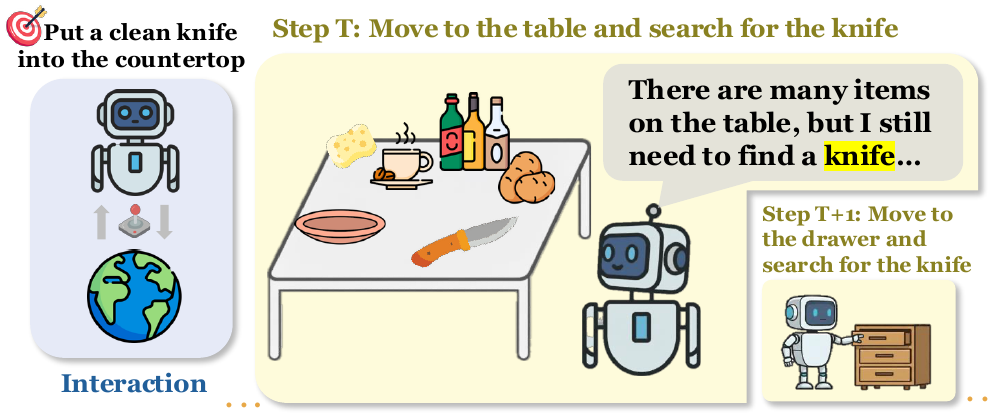}
    \caption{Illustrative example of \textit{observational neglect} in embodied agents. While the agent observes a knife on the target table, its subsequent internal belief (``I still need to find a knife'') fails to integrate the observed information, leading to an unnecessary search action.
    }
    \label{fig:intro-case}
    \vspace{-9pt}
\end{figure}

\begin{figure}[t!]
\centering
    \includegraphics[width=0.99\linewidth]{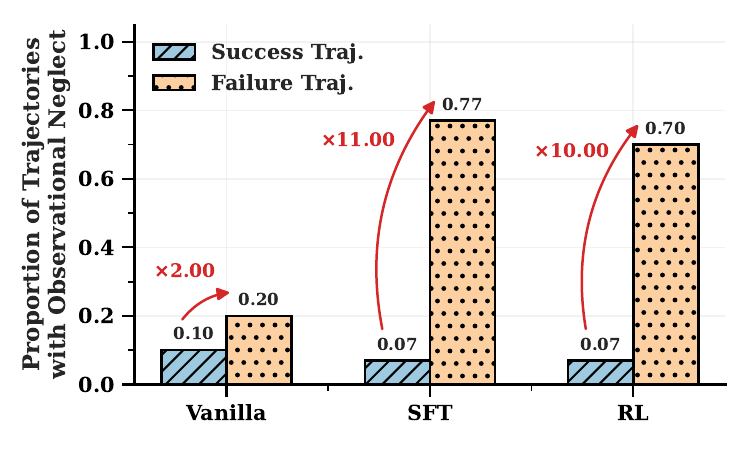}
    \caption{Statistical results of observational neglect on the ALFWorld benchmark.}
    \label{fig:intro-stat}
    \vspace{-9pt}
\end{figure}

However, we observe a significant gap between \textit{receiving} observations and effectively \textit{utilizing} them. 
As illustrated in Figure~\ref{fig:intro-case}, an agent may observe a knife on a countertop, yet its subsequent reasoning behaves as if the knife were still missing, initiating a redundant search action. 
We refer to this behavior as ``observational neglect'', where the agent observes but fails to integrate observed information into its internal reasoning process. 
Our statistical analysis (see Figure~\ref{fig:intro-stat}) on the ALFWorld~\cite{shridhar2020alfworld} benchmark reveals that such neglect is not a marginal error but a predominant failure mode in unsuccessful trajectories. 
Moreover, this behavior is widespread across various learning paradigms, from vanilla prompting to RL-tuned models, indicating a critical bottleneck in how embodied agents transform observations into their own beliefs, i.e., the internal understandings of the environment states.

To uncover the root cause of observational neglect, we conduct probing experiments to analyze the agent's belief dynamics (see Section~\ref{sec:pre_analyses}). 
Our analysis identifies a critical cognitive bias which we term \textbf{belief inertia}, a phenomenon that agents tend to stubbornly adhere to their prior expectations of action outcomes, even when faced with contradictory evidence. 
This inertia results in a belief-observation misalignment, where the agent's internal belief remains unchanged despite observing a changing environment. 
While recent studies have explored belief modeling~\cite{zhang2024agent, lidayan2025abbel}, they largely rely on implicit belief dynamics, where beliefs are updated passively and latently. 
Without an effective intervention when necessary, such strategies leave the agent's reasoning prone to being ``blinded'' by biased priors that overshadow its observations.

To address this, we advocate \textbf{active belief intervention}, shifting the paradigm from passive update to active cognitive management (see Section~\ref{sec:method}). 
We introduce the \textbf{E}stimate-\textbf{V}erify-\textbf{U}pdate (\textbf{\model}), a simple yet effective mechanism for belief intervention. 
Unlike previous works, \model decouples belief management from action generation by producing explicit belief states in a textual form. 
With \model, the agent first estimates an expected outcome, verifies it against actual observations through LLM-based reasoning, and finally updates its prior belief to a grounded posterior. 
Crucially, we integrate \model seamlessly into both prompting- and training-based agent learning methods.
Extensive experiments across diverse embodied benchmarks demonstrate that \model consistently yields substantial gains. 
Further in-depth analysis confirms that our \model mitigates belief inertia effectively.

In summary, our contributions are as follows:
\begin{itemize}[leftmargin=*, nolistsep]
    \item We identify and formalize belief inertia, a critical phenomenon in embodied agents where internal beliefs overshadow actual observations, leading to widespread observational neglect.
    \item We propose active belief intervention, implemented via the Estimate-Verify-Update (\model) mechanism. It provides a unified way to actively manage belief states and can be seamlessly integrated with various agent learning methods.
    \item We demonstrate the superiority and generalizability of \model through extensive experiments across multiple embodied benchmarks. Further analysis validates that \model significantly mitigates belief inertia, providing valuable insights into developing robust embodied agents.
\end{itemize}
\section{Preliminary}
\label{sec:preliminary}

\paragraph{Problem Formulation.}
The reasoning process of embodied agents is often formulated as a Partially Observable Markov Decision Process (POMDP), denoted as $(\mathcal{U}, \mathcal{S}, \mathcal{A}, \mathcal{O}, \mathcal{T}, \mathcal{Z}, \mathcal{R})$,
where $\mathcal{U}$ is the instruction space, $\mathcal{S}$ the hidden state space, $\mathcal{A}$ the action space, $\mathcal{O}$ the observation space, $\mathcal{T}$ the transition function, $\mathcal{Z}$ the observation function, and $\mathcal{R}$ the reward function. 
To isolate the cognitive aspects of agent–environment interaction from low-level perception, we focus on text-only settings where $\mathcal{U}$, $\mathcal{A}$, and $\mathcal{O}$ are all expressed in natural language.
Accordingly, we model an embodied agent as an LLM policy $\pi_\theta$ that generates textual actions.

\begin{figure}[ht]
    \centering
    \includegraphics[width=0.98\linewidth]{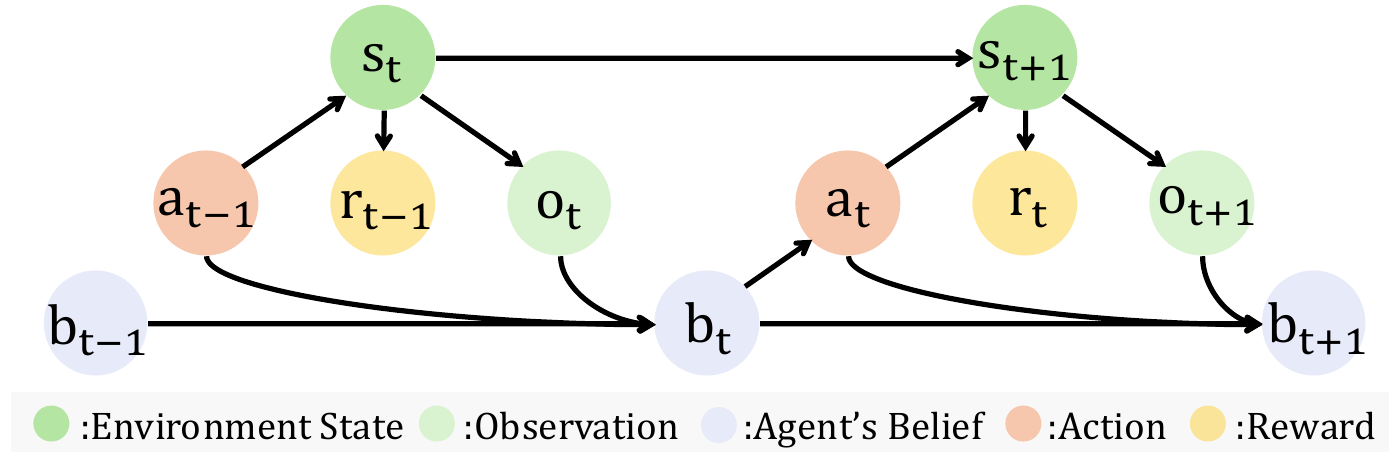}
    \caption{POMDP formulation in embodied agents.}
    \label{fig:pomdp}
    \vspace{-8pt}
\end{figure}

\paragraph{Agent Beliefs.}

In a POMDP, the true environment state $s_t$ is never observed directly, so the agent must maintain an internal belief state $b_t$ that summarizes its estimate of $s_t$ and serves as the basis for decision making.
As shown in Figure~\ref{fig:pomdp}, at each step the agent integrates new observation $o_t$ with its prior belief $b_{t-1}$ to obtain an updated belief $b_{t}$, and then reasons the next action conditioned on $b_{t}$, which in turn shapes future observations and rewards.
In many LLM-based agents, such belief dynamics are handled implicitly.
A common practice, i.e., ReAct-style~\cite{yao2022react} agent, appends all past actions and observations into an interaction history $h_t = (u, o_1, a_1, o_2, \ldots, o_t)$ and relies on LLMs to latently infer $b_t$ when reasoning the next action, without explicitly representing the belief state.
\section{Belief Inertia in Embodied Agents}
\label{sec:pre_analyses}

In this section, we investigate why ReAct-style embodied agents exhibit \emph{observational neglect}. 
Since this phenomenon manifests as a misalignment between the agent's implicit beliefs and the external environment, we employ a probing-based method to explicitly elicit and track these beliefs.

\subsection{Probing Agent Beliefs}
\label{sec:prob}

To probe an agent's beliefs, we append probing questions to the interaction history and utilize the agent's responses to decode its internal understanding of the environment. 

Specifically, given the interaction history $h_t$ at time $t$, we define a set of task-relevant environment variables $\mathcal{V}$, such as whether an object has been acquired or whether it is currently inside a container. 
For each variable $v \in \mathcal{V}$, we construct a corresponding yes--no probe question $q_v$ (e.g., ``Is the key currently in the box?''). 
We then construct a probing prompt by concatenating $h_t$ and $q_v$ and feeding it into the agent policy. Finally, we read the first-token logits for the candidate answers ``yes'' and ``no'', denoted by $\ell_{\mathsf{yes}}(h_t, q_v)$ and $\ell_{\mathsf{no}}(h_t, q_v)$.
We then define the raw belief value, the question-induced bias, and the debiased belief value as
\begin{equation}
\begin{aligned}
s(h_t, q_v)
&= \ell_{\mathsf{yes}}(h_t, q_v) - \ell_{\mathsf{no}}(h_t, q_v), \\
b(q_v)
&= \ell_{\mathsf{yes}}(h_\emptyset, q_v) - \ell_{\mathsf{no}}(h_\emptyset, q_v), \\
\beta(h_t, q_v)
&= s(h_t, q_v) - b(q_v).
\end{aligned}
\label{eq:belief-score}
\end{equation}
A positive $\beta(h_t, q_v)$ indicates an inclination to answer ``yes'' for $v$ under $h_t$ after filtering out question-induced bias, while a negative value indicates an inclination to answer ``no''.

For each variable $v$, let $y_v \in \{+1, -1\}$ denote the ground-truth answer in the current environment state, where $y_v = +1$ corresponds to ``yes'' and $y_v = -1$ to ``no''. We define the \textbf{True Belief Value}
\begin{equation}
  A(h_t, v)
  \;=\;
  y_v \, \beta(h_t, q_v),
  \label{eq:alignment-score}
\end{equation}
whose sign indicates whether the probed belief agrees with the true state ($A(h_t, v) > 0$) or not ($A(h_t, v) < 0$), and whose magnitude $|A(h_t, v)|$ serves as a proxy for the confidence of this belief.
Correlation analyses between probing results and the agent's behavior support the reliability of our probing method (see Appendix~\ref{app:probing_reliability}).

\begin{figure}[t!]
    \centering
    \includegraphics[width=0.98\linewidth]{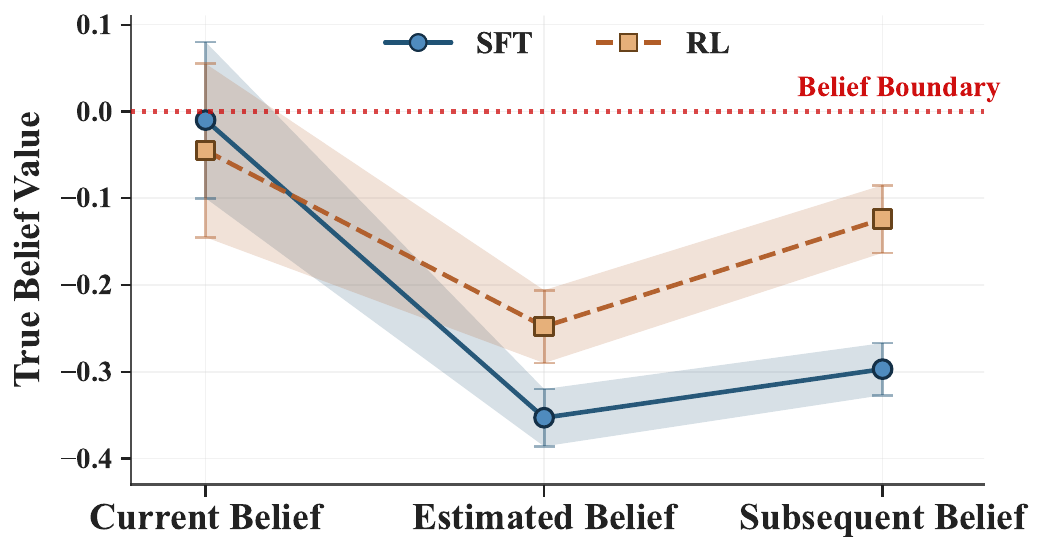}
    \caption{Probing results of belief dynamics across three stages. The belief boundary separates positive (correct) beliefs from negative (incorrect) ones.}
    \label{fig:pre_ana3}
    \vspace{-8pt}
\end{figure}

\subsection{Belief Inertia Phenomenon}
\label{pre_ana: belief_inertia_dynamics}

To analyze the agent's decision-making process when the \emph{observational neglect} occurs, we apply the above probing method to examine how its beliefs are evolved.

We first collect 100 observational-neglect cases from both SFT-trained and RL-trained agents in ALFWorld~\cite{shridhar2020alfworld}, where the agent’s internal reasoning for the next action $a_{t}$ neglects the newly received observation $o_t$.
We then extract $o_t$ from $v$ that captures the critical feedback and construct the corresponding probe question $q_v$.
To examine belief updating for this variable, we analyze how the agent’s beliefs evolve between the previous action $a_{t-1}$ that produces the new observation $o_t$ and the subsequent action $a_t$.
Specifically, we probe the agent's belief about $v$ at three stages:
(1) \textbf{Current belief}, representing the belief before taking $a_{t-1}$, probed under $(u, o_1, a_1, o_2, \ldots, \bm{o_{t-1}})$;
(2) \textbf{Estimated belief}, representing the belief after taking $a_{t-1}$ but before observing $o_t$, probed under $(u, o_1, a_1, o_2, \ldots, \bm{o_{t-1}, a_{t-1}})$;
and (3) \textbf{Subsequent belief}, representing the belief after receiving $o_t$, probed under $(u, o_1, a_1, o_2, \ldots, \bm{o_{t-1}, a_{t-1}, o_t})$.

Figure~\ref{fig:pre_ana3} visualizes the probed True Belief Value across the three stages for both SFT- and RL-trained agents. At the initial stage, $A(c_t, v)$ lies near the belief boundary, indicating that the agent does not hold a strong prior about the query. 
After taking an action, the agent forms a strong but incorrect belief about $v$. Although the subsequent belief increases slightly after receiving $o_t$, it remains negative for both agent types. This persistence suggests that the new environmental feedback fails to update the agent's internal state. 
Consequently, the agent reasons the next action based on a stale belief formed immediately after the previous action, which remains inconsistent with the environment state. We term this failure to update \textbf{belief inertia}, a phenomenon that manifests as observational neglect during subsequent reasoning.

\subsection{Belief Intervention}
\label{sec:belief_intervention}

\begin{figure}[t!]
  \includegraphics[width=0.98\linewidth]{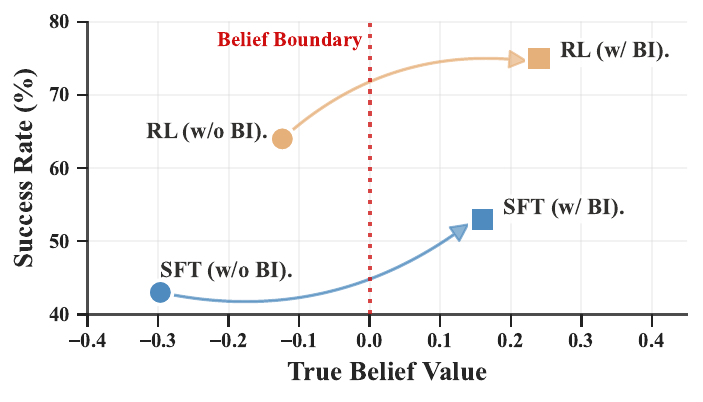}
  \caption{Impact of oracle belief intervention (BI).}
  \label{fig:belief_interven}
  \vspace{-8pt}
\end{figure}

To examine whether the belief inertial causes observational neglect, we conduct a \textbf{Belief Intervention (BI)} experiment: we manually correct the agent’s belief to match the true environment state and then observe whether it can reason and act correctly.

Specifically, we adopt the same observational-neglect cases as mentioned above and evaluate both the SFT-trained and RL-trained agents under two distinct settings.
In the typical setting (\textbf{w/o BI}), the agent generates $a_{t}$ conditioned on the standard interaction history $h_t$.
In the intervention setting (\textbf{w/ BI}), we explicitly append a description of the oracle environment state $s_t^{*}$ to the interaction history, yielding $(u, o_1, a_1, \ldots, o_t, s_t^{*})$.
In both settings, we apply our probing method to assess the agent's belief about the environment variables relevant to $o_t$ and report the task success rate on these observational-neglect cases.

As illustrated in Figure~\ref{fig:belief_interven}, belief intervention shifts the agents' internal states across the Belief Boundary, correlating directly with improved task performance.
Without intervention, both SFT-trained and RL-trained agents linger in the negative True Belief Value region, indicating a persistence of incorrect beliefs that corresponds to lower success rates.
Upon intervention, the True Belief Value becomes positive, signifying the adoption of the correct environmental state.
Crucially, this belief correction translates into a marked increase in success rates for both agents.
These results confirm that the primary bottleneck in observational neglect cases is the failure to update beliefs. By manually aligning the belief state with the actual environment, we mitigate the downstream consequences of this failure, demonstrating that the agents possess the necessary reasoning capabilities to succeed once the belief barrier is removed.
\section{Method: Active Belief Intervention}
\label{sec:method}

\begin{figure*}[t!]
    \centering
    \includegraphics[width=0.95\textwidth]{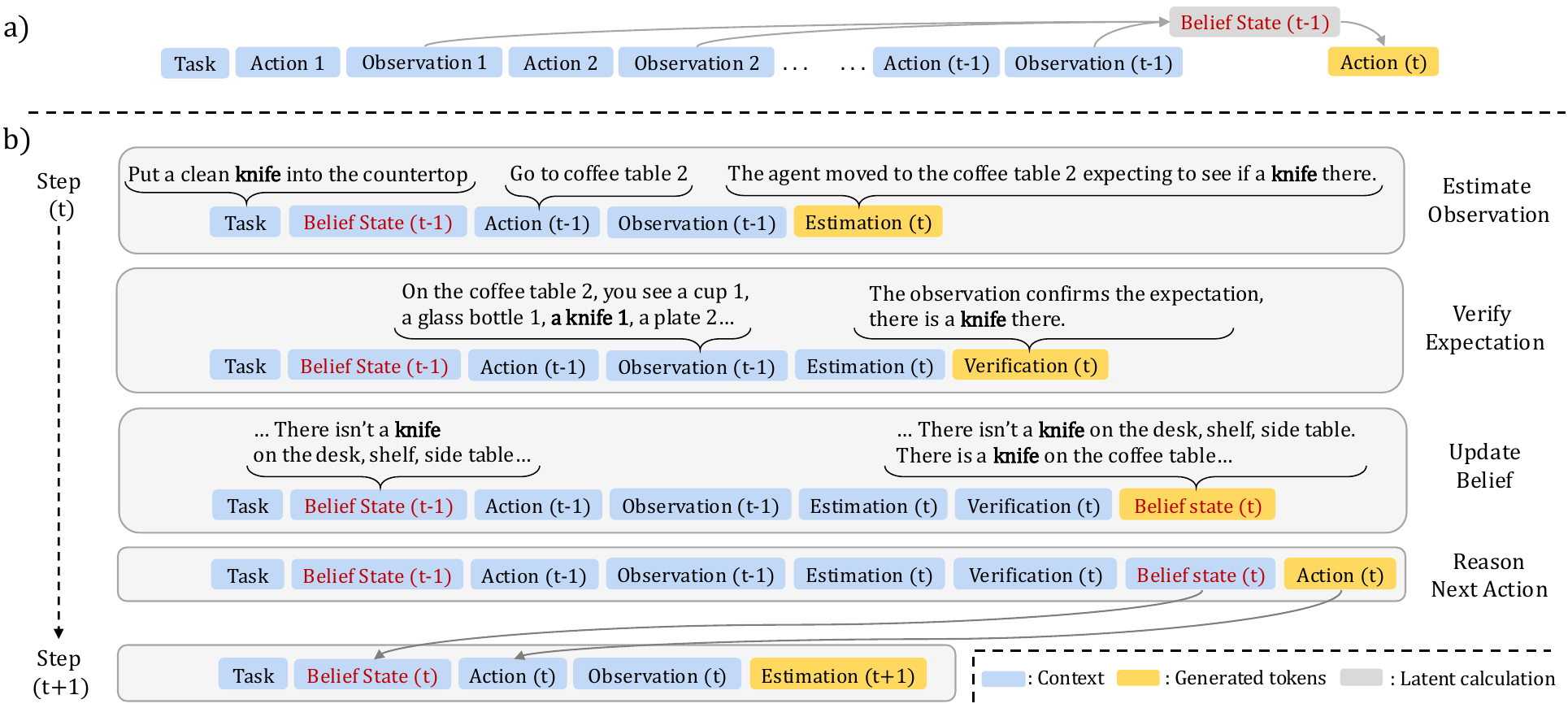}
    \caption{Overview of our proposed active belief intervention method. Compared to typical belief modeling methods (\textit{top}), we introduce a unified Estimate-Verify-Update (\model) mechanism (\textit{bottom}).}
    \label{fig:overview}
    \vspace{-9pt}
\end{figure*}

Drawing upon the crucial findings in Section~\ref{sec:pre_analyses}, we advocate active belief intervention and introduce a simple yet effective \textbf{E}stimate-\textbf{V}erify-\textbf{U}pdate (\textbf{\model}) mechanism that estimates, verifies, and updates beliefs actively through a unified perspective.
Figure~\ref{fig:overview} shows the overview of our approach.

\subsection{Estimate-Verify-Update Mechanism}
\label{sec:method_dynamics}

In contrast to typical ReAct-style agents, which passively encode the entire interaction history as an implicit belief about the environment within the latent model parameters, our \model mechanism maintains an explicit belief state $B_t$, a natural language summary that sufficiently represents the understanding of the environment. 
Crucially, \model recursively takes the previous belief state as input and evolves it through a structured loop of estimation, verification, and update by the agent itself.

\paragraph{Estimation.}
Initially, the agent attempts to predict the immediate consequence of its previous action before processing the actual new observation. 
In this step, the agent establishes a baseline expectation by estimating action outcomes $E_t$ as:
\begin{equation}
E_t \sim \pi_{\theta}(\cdot \mid B_{t-1},a_{t-1},o_{t}),
\end{equation}
where $E_t$ describes what the agent expects to observe, explicitly modeling its expectation.

\paragraph{Verification.}
The agent then processes the actual observation $o_t$ from the environment. Instead of updating the belief directly, the agent first generates a verification evidence $V_t$ to compare its estimation against the actual observation:
\begin{equation}
V_t \sim \pi_{\theta}(\cdot \mid B_{t-1},a_{t-1},o_{t},E_t).
\end{equation}
Here, $V_t$ serves as a structured ``surprise signal'' that explicitly captures whether the observation confirms or contradicts the expectation, preventing the agent from hallucinating success or overlooking contradictory evidence.

\paragraph{Belief Update.}

Finally, the agent synthesizes the reasoning chain to transition from the previous belief state $B_{t-1}$ to the current belief state $B_t$. This process takes the prior belief, the initial estimation $E_t$, and the verification evidence $V_t$ (i.e., the surprise signal) as inputs:
\begin{equation}
B_t \sim \pi_\theta(\cdot \mid B_{t-1}, a_{t-1}, o_t, E_t, V_t).
\end{equation}
By leveraging this surprise-aware verification $V_t$, the model ensures that the new belief state $B_t$ (e.g., “There is a knife on the coffee table…”) accurately reflects the latest environmental changes while retaining valid historical information about the environment. This updated belief then serves as the foundation for subsequent reasoning.

\subsection{A Unified Intervention Perspective}
\label{sec:method_paradigms}

Our \model is a general mechanism that decouples \textit{state maintenance} from \textit{action generation}. 
This separation allows for a unified formulation that is agnostic to both the prompting-based methods and the training-based algorithms.

\subsubsection{Prompting-based Belief Intervention}

Standard prompting-based methods typically employ a prompting strategy $\mathcal{S}$ to directly map interaction history $h_t$ to an action $a_t$.
In our approach, we realize belief intervention by augmenting the original strategy $\mathcal{S}$ with specific instructions designed to enable active belief dynamics. We denote this belief-enhanced strategy as $\mathcal{S}^*$, which explicitly guides the agent to perform active belief intervention before decision-making. This process is formulated as:
\begin{equation}
    (E_t, V_t, B_t, a_t) \sim \pi_\theta\left(\cdot \mid \mathcal{S}^*(B_{t-1}, a_{t-1}, o_t)\right).
\end{equation}

By requiring the agent to explicitly model its belief dynamics prior to the action $a_t$, 
this intervention ensures that the agent's decision-making is grounded in a structured and updated understanding of the environment, rather than relying solely on implicit patterns within the raw history.

\begin{table*}[t!]
\centering
\resizebox{1.0\textwidth}{!}{
\begin{tabular}{llcccccccccccc}
\toprule
\multirow{2}{*}{\textbf{}} &
\multirow{2}{*}{\textbf{Method}} &
\multicolumn{4}{c}{\textbf{ALFWorld}} &
\multicolumn{4}{c}{\textbf{VirtualHome}} &
\multicolumn{4}{c}{\textbf{ScienceWorld}} \\
\cmidrule(lr){3-6} \cmidrule(lr){7-10} \cmidrule(lr){11-14}
& & \multicolumn{2}{c}{Seen} & \multicolumn{2}{c}{Unseen} &
  \multicolumn{2}{c}{Seen} & \multicolumn{2}{c}{Unseen} &
  \multicolumn{2}{c}{Seen} & \multicolumn{2}{c}{Unseen} \\
\midrule

\multirow{7}{*}{\shortstack[l]{Prompting}} 
& \multicolumn{13}{c}{\cellcolor{modelbg} \texttt{DeepSeek V3.2}} \\
& NoThinking & \base{50.7} & \n & \base{42.3} & \n & \base{8.0} & \n & \base{7.2} & \n & \base{47.0} & \n & \base{46.0} & \n \\
& $\hookrightarrow$ w/ \model (Ours) &
    \our{55.0} & \textbf{\gain{4.3}} &
    \our{47.6} & \textbf{\gain{4.3}} &
    \our{12.8} & \textbf{\gain{4.8}} &
    \our{12.8} & \textbf{\gain{5.6}} &
    \our{55.0} & \textbf{\gain{8.0}} &
    \our{52.2} & \textbf{\gain{6.2}} \\
& Plan-and-Act & \base{52.1} & \n & \base{44.8} & \n & \base{12.8} & \n & \base{12.8} & \n & \base{55.0} & \n & \base{50.9} & \n \\
& $\hookrightarrow$ w/ \model (Ours) &
    \our{53.6} & \gain{1.5} &
    \our{46.3} & \gain{1.5} &
    \our{15.2} & \gain{2.4} &
    \textbf{\our{14.4}} & \gain{1.6} &
    \our{58.3} & \gain{3.3} &
    \our{55.3} & \gain{4.4} \\
& ReAct & \base{55.7} & \n & \base{47.6} & \n & \base{13.6} & \n & \base{12.8} & \n & \base{60.3} & \n & \base{57.8} & \n \\
& $\hookrightarrow$ w/ \model (Ours) &
    \textbf{\our{56.4}} & \gain{0.7} &
    \textbf{\our{49.8}} & \gain{2.2} &
    \textbf{\our{16.0}} & \gain{2.4} &
    \our{13.6} & \gain{0.8} &
    \textbf{\our{62.3}} & \gain{2.0} &
    \textbf{\our{60.9}} & \gain{3.1} \\
\midrule

\multirow{14}{*}{\shortstack[l]{Training}} 
& \multicolumn{13}{c}{\cellcolor{modelbg}\texttt{Qwen3-1.7B-Instruct}} \\
& SFT & \base{37.1} & \n & \base{20.1} & \n & \base{7.2} & \n & \base{8.0} & \n & \base{7.3} & \n & \base{11.2} & \n \\
& $\hookrightarrow$ w/ \model (Ours) &
    \our{41.4} & \gain{4.3} &
    \our{33.6} & \textbf{\gain{13.5}} &
    \our{16.0} & \textbf{\gain{8.8}} &
    \our{25.6} & \textbf{\gain{17.6}} &
    \our{23.2} & \gain{15.9} &
    \our{24.8} & \textbf{\gain{13.6}} \\
& PPO & \base{42.1} & \n & \base{32.0} & \n & \base{10.4} & \n & \base{22.4} & \n & \base{37.0} & \n & \base{41.0} & \n \\
& $\hookrightarrow$ w/ \model (Ours) &
    \our{47.1} & \gain{5.0} &
    \our{40.3} & \gain{8.3} &
    \our{17.6} & \gain{7.2} &
    \our{28.8} & \gain{6.4} &
    \our{62.9} & \textbf{\gain{25.9}} &
    \textbf{\our{54.0}} & \gain{13.0} \\
& GRPO & \base{47.0} & \n & \base{44.0} & \n & \base{15.7} & \n & \base{19.4} & \n & \base{41.7} & \n & \base{42.9} & \n \\
& $\hookrightarrow$ w/ \model (Ours) &
    \textbf{\our{52.1}} & \textbf{\gain{5.1}} &
    \textbf{\our{49.3}} & \gain{5.3} &
    \textbf{\our{20.0}} & \gain{4.3} &
    \textbf{\our{36.9}} & \gain{17.5} &
    \textbf{\our{47.7}} & \gain{6.0} &
    \our{50.3} & \gain{7.4} \\ 

\cmidrule(l){2-14} 

& \multicolumn{13}{c}{\cellcolor{modelbg}\texttt{Qwen2.5-3B-Instruct}} \\
& SFT & \base{65.7} & \n & \base{50.7} & \n & \base{20.0} & \n & \base{20.0} & \n & \base{19.9} & \n & \base{13.7} & \n \\
& $\hookrightarrow$ w/ \model (Ours) &
    \our{70.0} & \textbf{\gain{4.3}} &
    \our{56.7} & \gain{6.0} &
    \our{27.2} & \textbf{\gain{7.2}} &
    \our{34.4} & \textbf{\gain{14.4}} &
    \our{49.0} & \textbf{\gain{29.1}} &
    \our{45.3} & \textbf{\gain{31.6}} \\
& PPO & \base{77.8} & \n & \base{54.4} & \n & \base{24.0} & \n & \base{23.2} & \n & \base{53.6} & \n & \base{51.6} & \n \\
& $\hookrightarrow$ w/ \model (Ours) &
    \our{79.3} & \gain{1.5} &
    \our{58.2} & \gain{3.8} &
    \our{28.8} & \gain{4.8} &
    \our{35.2} & \gain{12.0} &
    \our{60.3} & \gain{6.7} &
    \our{62.7} & \gain{11.1} \\
& GRPO & \base{83.6} & \n & \base{70.8} & \n & \base{25.6} & \n & \base{24.8} & \n & \base{49.7} & \n & \base{52.8} & \n \\
& $\hookrightarrow$ w/ \model (Ours) &
    \textbf{\our{85.7}} & \gain{2.1} &
    \textbf{\our{79.1}} & \textbf{\gain{8.3}} &
    \textbf{\our{31.2}} & \gain{5.6} &
    \textbf{\our{36.0}} & \gain{11.2} &
    \textbf{\our{70.9}} & \gain{10.6} &
    \textbf{\our{70.8}} & \gain{18.0} \\
\bottomrule
\end{tabular}
}
\caption{Main results of success rates (\%) on three representative embodied benchmarks. ``Seen'' and ``Unseen'' denote held-out test sets with tasks seen and unseen during training, respectively. ``w/ \model'' denotes plugging our \model mechanism into base methods. \textbf{Bold} values represent the best performance within each backbone model group.}
\label{tab:main_results}
\vspace{-8pt}
\end{table*}

\begin{figure*}[t!]
    \centering
    \includegraphics[width=0.95\textwidth]{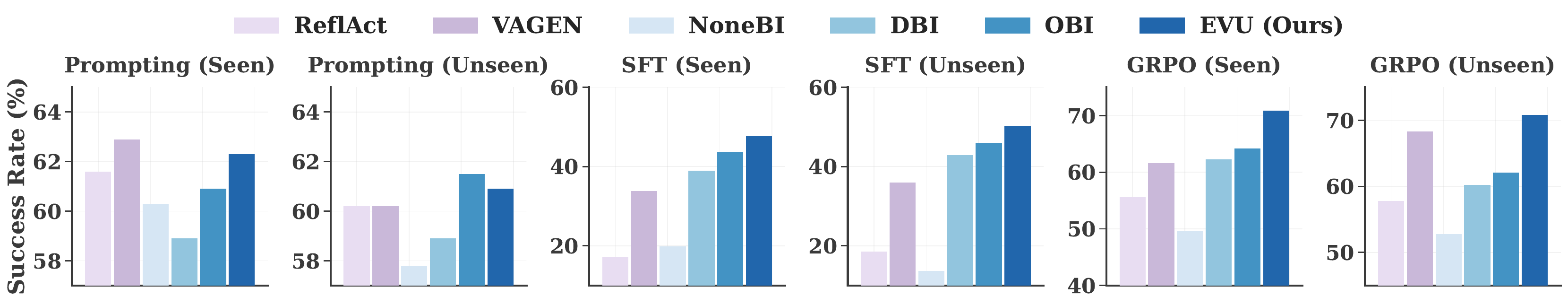}
    \caption{Success rates (\%) of different methods with belief intervention variants.}
    \label{fig:ablation}
    \vspace{-9pt}
\end{figure*}

\subsubsection{Training-based Belief Intervention}

In training-based approaches, we update the model parameters $\theta$ to internalize the belief update mechanism.
Unlike typical training methods that focus solely on optimizing action generation, we unify belief update and action generation into an autoregressive process and optimize them jointly.

Formally, at each time step $t$, the model takes the previous belief state $B_{t-1}$ and the recent interaction history $(a_{t-1}, o_{t-1})$ as inputs to generate the current reasoning chain and action:
\begin{equation}
(E_t, V_t, B_t, a_t) \sim \pi_\theta(\cdot \mid B_{t-1}, a_{t-1}, o_t).
\end{equation}

To optimize this process, we define a general objective function $\mathcal{J}(\theta)$, which represents the expected utility of the generated trajectory. Depending on the training paradigm, $\mathcal{J}(\theta)$ can be flexibly instantiated as the negative log-likelihood in Supervised Fine-Tuning (SFT) or the expected reward in Reinforcement Learning (e.g., PPO, GRPO). 
This is formulated as:
\begin{equation}
\theta^* = \operatorname*{argmax}_\theta \mathcal{J}(\theta) = \operatorname*{argmax}_\theta \mathbb{E}_{\tau \sim \pi_\theta} [U(\tau)],
\end{equation}
where $\tau = (B_0, a_0, o_0, E_1, V_1, B_1, a_1, \dots)$ represents the augmented trajectory containing both cognitive states and external actions. By maximizing $\mathcal{J}(\theta)$, the optimization algorithm adjusts the probability mass not just for the final action $a_t$, but for the entire reasoning chain ($E_t, V_t, B_t$). This ensures that the model learns to maintain high-quality beliefs that causally lead to optimal actions, allowing gradients (or reward signals) to propagate through the belief update process.

\section{Experiments}
\label{sec:exp}

\subsection{Experimental Setup}

\paragraph{Benchmarks.}
We evaluate our method on three representative embodied agent benchmarks: ALFWorld~\citep{shridhar2020alfworld}, VirtualHome~\citep{puig2018virtualhome}, and ScienceWorld~\citep{wang2022scienceworld}.
Following prior studies~\citep{song2024trial,wang2025spa}, we adopt Success Rate (SR) as our primary evaluation metric and evaluate agents on both seen and unseen scenarios.
Appendix~\ref{app:datasets} provides more details of these datasets.

\paragraph{Baseline Methods.}

We evaluate our method by integrating it into two categories of baselines and measuring the resulting performance gains: (1) prompting-based methods, including No-Thinking~\citep{ma2025reasoning}, Plan-and-Act~\citep{kim2025reflact}, and ReAct~\citep{yao2022react}; and (2) training-based methods, including SFT~\citep{chen2023fireact}, PPO~\citep{schulman2017proximal}, and GRPO~\citep{shao2024deepseekmath}. Additional details about these baselines are provided in Appendix~\ref{app:baseline}.

\paragraph{Implementation Details.}
We conduct experiments on DeepSeek V3.2~\citep{liu2025deepseek} for prompting-based evaluations, as well as Qwen2.5-3B-Instruct~\citep{yang2025qwen3} and Qwen3-1.7B-Instruct~\citep{yang2025qwen3} for training-based evaluations.
For the SFT phase, the training epochs are set to 3.
For the RL phase, the training process consists of 250 steps, and we select the checkpoint with the best performance on the validation set for final testing.
During inference, the decoding temperature of the LLMs is set to 0.0 for deterministic generation.
Detailed hyperparameters and prompt designs are provided in the Appendix~\ref{app:more_implementation}.

\subsection{Main Results}

\begin{figure}[t]
    \centering
    \includegraphics[width=0.98\linewidth]{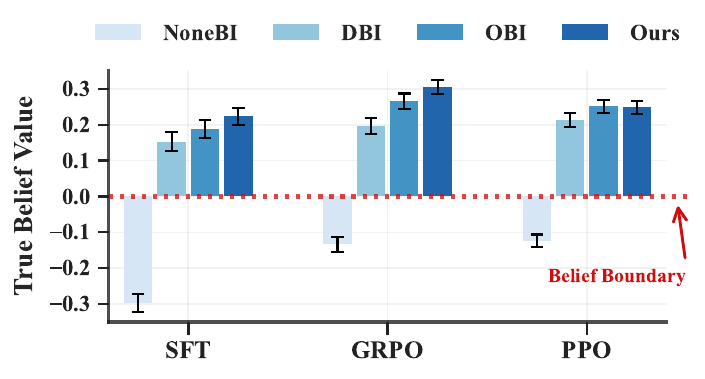}
    \caption{Quantitative probing results of different belief intervention methods in mitigating belief inertia.
    }
    \label{fig:IBU}
    \vspace{-8pt}
\end{figure}

Table~\ref{tab:main_results} presents a comprehensive evaluation of different methods across three benchmarks. 
We summarize key findings of \model below:

\paragraph{Consistent improvement across benchmarks and backbone models.}
As shown in Table~\ref{tab:main_results}, \model consistently outperforms all baseline methods across all three benchmarks, demonstrating its effectiveness and robustness.
Notably, we observe that the average performance gain on Unseen splits (+9.21) is higher than that on Seen splits (+6.8). This indicates that our method effectively grounds the agent even when facing novel observations in OOD scenarios, thereby substantially enhancing generalization capabilities.
Additional comparisons with advanced context-management and search-based agentic baselines are provided in Appendix~\ref{app:advanced_baselines}, where EVU remains consistently beneficial.

\paragraph{Robustness in both prompting- and training-based settings.}

Our method demonstrates remarkable flexibility by seamlessly integrating with both prompting-based and training-based methods. As shown in Table~\ref{tab:main_results}, \model consistently enhances performance across these distinct modes.
Notably, the average improvement in training-based settings (+10.37) significantly exceeds that in prompting-based settings (+3.28). This disparity suggests that while \model serves as an effective inference-time guidance, its full potential is unleashed when the backbone model is allowed to internalize the active belief update process, leading to more substantial performance gains.

\section{Analyses and Discussions}
\label{sec:analyses}

\subsection{Variant Analysis}

\begin{figure}[t]
    \centering
    \includegraphics[width=0.98\linewidth]{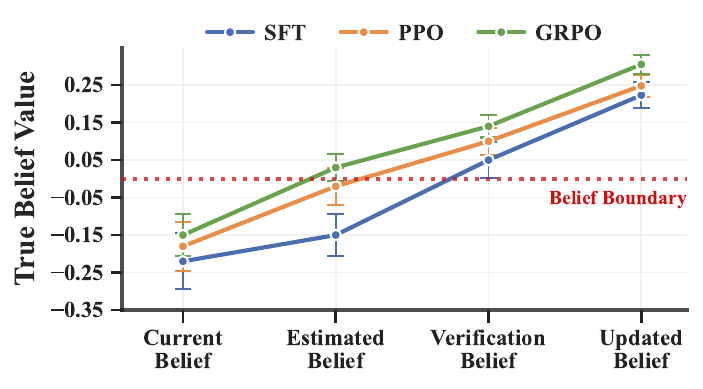}
    \caption{Quantitative probing results of different phase in mitigating belief inertia.
    }
    \label{fig:CHANGE_BELIEF}
    \vspace{-8pt}
\end{figure} 

To investigate the efficacy of belief intervention (BI), we examined distinct intervention strategies injected before the thought-action generation process. These include: \textbf{NoneBI}, which serves as the baseline without any intervention; \textbf{DBI} (Direct Belief Intervention), where the belief state is generated directly; \textbf{OBI} (Observation-based Belief Intervention), which compels the agent to reiterate the recent observation prior to forming a belief; \textbf{ReflAct}, which encourages the agent to reflect on its progress relative to the goal; and \textbf{VAGEN}~\cite{wang2025vagen}, which tasks the agent with predicting the environmental state following a potential action. Please refer to Appendix~\ref{app:variant_details} for more details.

The comparative results are presented in Figure~\ref{fig:ablation}. First, we observe that methods incorporating belief intervention consistently outperform the baseline (NoneBI) across the majority of settings. This trend underscores the fundamental efficacy of explicit belief modeling in enhancing task performance. Second, and more importantly, our proposed \model achieves superior performance compared to all other intervention variants across diverse paradigms and evaluation splits. This consistent dominance suggests that the \model mechanism provides a more robust and accurate strategy for belief generation than simple repetition or reflection, thereby serving as an effective intervention strategy.

\subsection{Analysis on Belief Inertia Mitigation}

\textbf{Can \model effectively mitigate the belief inertia phenomenon?} 
To investigate this, we conduct a dedicated analysis to examine how our method behaves on the observational neglect cases collected in Section~\ref{pre_ana: belief_inertia_dynamics}.
Specifically, we employ our probing method to detect the agent's belief immediately prior to the decision-making phase. (See Appendix~\ref{app:belief_intertia_mitigation} for detailed experimental configurations.) 
As illustrated in Figure~\ref{fig:IBU}, we observe that the true belief values for our method are consistently positive. This indicates that the agent's internal belief state aligns with the actual environmental state, successfully overcoming belief inertia. Furthermore, compared to other variants, \model exhibits the highest true belief values. 
This superiority demonstrates that our approach not only corrects the belief, but also achieves the highest level of confidence in the true state of the environment.

\noindent\textbf{How does \model take effect to mitigate the belief inertia?} To understand the internal mechanism, we conduct further analysis to observe the agent's belief dynamic evolution using the same set of observational neglect cases mentioned above. We probe the agent's belief at different stages within our \model framework. Please refer to the Appendix~\ref{app:belief_intertia_mitigation} for more details. 
As illustrated in Figure~\ref{fig:CHANGE_BELIEF}, the agent's belief value starts in the negative region and gradually ascends. While the estimation and verification phases push the belief towards and across the boundary, respectively, it is the final update phase that significantly boosts the value to fully align with the ground truth. This indicates that the update stage is the decisive factor for synchronization, effectively building upon the foundations laid by the preceding phases.

\begin{figure}[t!]
    \centering
    \includegraphics[width=0.98\linewidth]{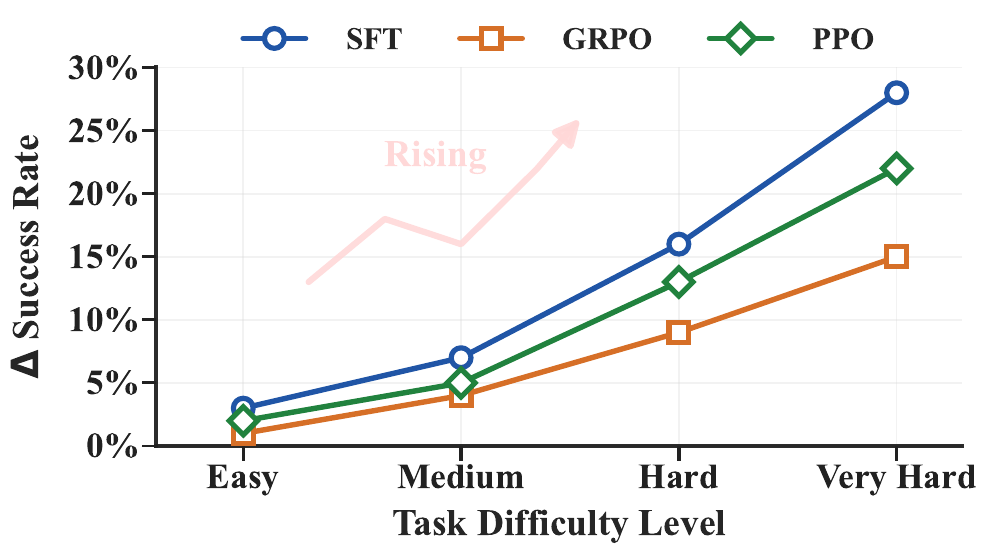}
    \caption{Relative improvement of our method compared to baselines (SFT, GRPO, PPO) across different levels of task difficulty.}
    \label{fig:BSE}
    \vspace{-8pt}
\end{figure}

\subsection{Impact on Task Difficulty}
\label{sec:impact_task_difficulty}

We further investigate the necessity of active belief intervention as the task difficulty increases. Difficult tasks inherently involve longer interaction horizons, requiring the agent to process more observations. 
Consequently, the ability to maintain accurate and synchronized belief states becomes increasingly critical during these interactions. 
To verify this, we evaluate the relative success rate improvement ($\Delta$ Success Rate) of our method compared to three training baselines (SFT, GRPO, and PPO) across four difficulty levels (see Appendix~\ref{app:task_difficulty} for details).

As shown in Figure~\ref{fig:BSE}, our approach consistently outperforms the baselines across all settings. More importantly, the performance gap exhibits a clear rising trend: as the difficulty escalates from ``Easy'' to ``Very Hard,'' the relative improvement becomes significantly more pronounced. 
This validates that active belief intervention enables the agent to understand the environment more deeply, thereby preventing compounding errors.

\section{Analysis on Computational Overhead}
\label{sec:ana_co}

\begin{figure}[t!]
    \centering
    \includegraphics[width=0.98\linewidth]{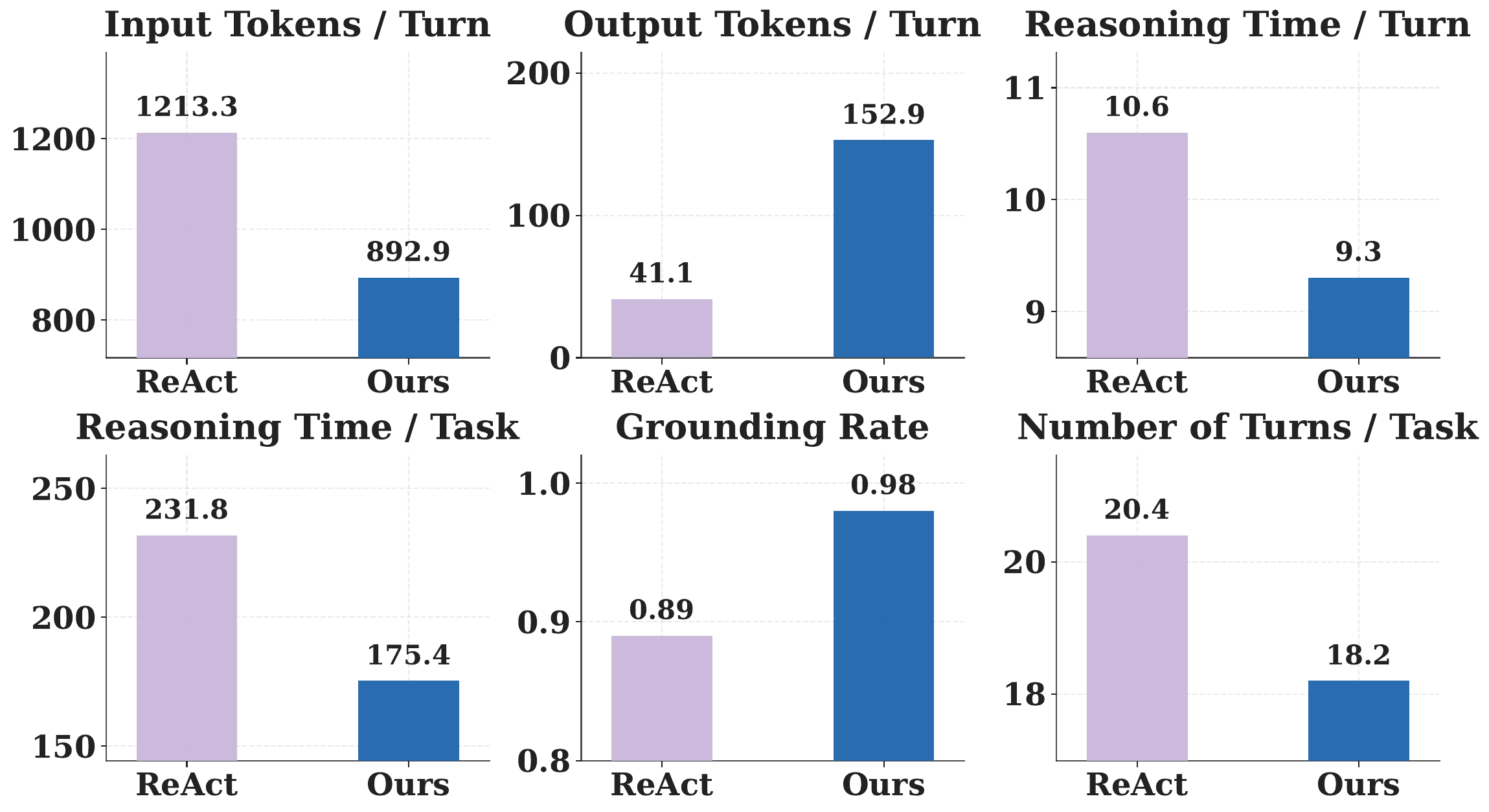}
    \caption{Comparison between ReAct and Ours in terms of computational overhead.}
    \label{fig:control_exp}
    \vspace{-8pt}
\end{figure} 

\textbf{Do the performance gains of EVU come with substantial additional computational overhead?}
To investigate this, we compare the standard ReAct baseline and ReAct+EVU on the Qwen2.5-3B backbone. We report the average input/output tokens, reasoning latency, grounding rate, and average number of turns per task.

As illustrated in Figure~\ref{fig:control_exp}, although EVU naturally increases the number of output tokens per turn due to the generation of the \textit{Estimate}, \textit{Verify}, and \textit{Update} components, it substantially reduces the input context length by compressing the verbose interaction history into a concise belief state. As a result, the average total token consumption per task decreases from 1213.3 to 892.9, while the average reasoning latency per task also decreases from 231.8 to 175.4. In addition, EVU improves the grounding rate from 0.89 to 0.98 and reduces the average number of turns from 20.4 to 18.2, indicating that the agent avoids invalid exploration and plans more effectively. These results show that EVU does not incur a larger overall computational burden; instead, it improves efficiency by replacing low-quality exploration with more grounded and directed reasoning.

\noindent
\textbf{Are the gains of EVU simply due to increased inference computation (test-time scaling)?} We further conducted a controlled experiment on ALFWorld with DeepSeek V3.2 in the prompting-based setting. Specifically, we compare EVU against a \textit{Multiple Reflection} baseline, where the agent is prompted to reflect multiple times to deliberately increase the reasoning token budget. This setting allows us to examine whether the gain of EVU comes merely from consuming more reasoning tokens, or from the specific structure of the EVU process itself.

As shown in Table~\ref{tab:control_scaling}, EVU achieves a higher success rate than \textit{Multiple Reflection} while using fewer reasoning tokens, which first rules out the explanation that the gain simply comes from spending more computation at inference time. More importantly, this comparison also highlights a key distinction between EVU and standard reflection. \textit{Multiple Reflection} is retrospective: it asks the agent to reconsider whether it may have made a mistake after the fact. In contrast, EVU is predictive and discrepancy-driven. 
The \textit{Estimate} step makes the agent explicitly predict the expected outcome before observing the new state, giving the \textit{Verify} step a concrete reference point. This allows EVU to detect a ``surprise signal,'' i.e., the mismatch between expectation and observation. The \textit{Update} step then revises the belief state accordingly, rather than merely triggering another round of generic self-correction.
This predictive mechanism is particularly important for overcoming belief inertia: without an explicit prior estimate, standard reflection may remain trapped in the agent’s previous belief and fail to recognize the reality gap. Therefore, EVU improves performance not by encouraging longer reasoning chains, but by enforcing a simple yet effective belief-correction process that is more targeted than standard reflection.

\begin{table}[t]
\centering
\small
\begin{tabular}{lcc}
\toprule
Method & \# Reasoning Tokens & SR (\%) \\
\midrule
Multiple Reflection & 153.3 & 46.3 \\
EVU (Ours) & 138.0 & 49.8 \\
\bottomrule
\end{tabular}
\caption{Comparison between our \model and the Multiple Reflection method.}
\label{tab:control_scaling}
\vspace{-6pt}
\end{table}

\section{Related Work}
\label{sec:related_work}

\paragraph{Embodied Planning.}
Recent advancements in Large Language Models (LLMs) have empowered embodied agents to engage in complex embodied planning~\citep{li-etal-2025-foundation,yang2025embodiedbench,liao2025think}.
To facilitate effective decision-making, existing studies employ diverse strategies: Prompting methods~\cite{yao2022react,shinn2023reflexion,yao2023tree} structure reasoning at inference time, supervised finetuning~\citep{chen2023fireact,wang2025steca,qiao2024agent} internalize expert priors, and reinforcement learning refines policy from reward signals~\citep{song2024trial,chen2025era,wang2025spa,zhang2025embodied}.  
However, these methods primarily prioritize action optimization, often neglecting the critical need to maintain a reliable internal world model amidst dynamic environmental changes~\citep{huang2023grounded,wang2024e2cl,kim2025reflact}.

\paragraph{Agent Beliefs.}
To achieve goals, agents are required to maintain and update their internal beliefs during interaction.
Prior work has highlighted multiple complementary facets of such beliefs.
First, agents often rely on task belief—estimates of progress and knowledge—to support long-horizon planning~\citep{qiao2024agent,wang2024e2cl,zhang2024agent}.
Second, in human-agent interaction, agents capture users’ latent intent to interpret ambiguous instructions~\citep{lin2025ask,ramrakhya2025grounding}.
Third, in cooperative multi-agent settings, agents track others’ capabilities, objectives, and likely future actions to enable coordination~\citep{fan2025somi,licua2024mindforge, wang2026foresightoptimizationstrategicreasoning}.
In this work, we focus on agent beliefs regarding the evolving environment, which grounds reasoning under environment state changes during interactions.
\section{Conclusion}

In this work, we identify and formalize \emph{belief inertia} as a key failure mode of LLM-based embodied agents, where they stubbornly adhere to prior beliefs despite explicit observations. To address this issue, we advocate active belief intervention and instantiate it with the Estimate-Verify-Update (\model) mechanism. By integrating \model into both prompting-based and training-based methods, we mitigate belief inertia effectively, thereby obtaining consistent improvements in task performance across multiple embodied agent benchmarks.

\newpage

\section*{Limitations}

While our approach demonstrates superior performance compared to baseline methods, it is important to acknowledge the limitations of our current work as follows:

(1) Dependency on Observation Quality: Our method relies on the quality and granularity of environmental observations to update its belief dynamics. In scenarios with extremely sparse, noisy, or ambiguous feedback, where the ground truth is difficult to discern even with active reasoning, the agent's belief updates may become unstable. Future work could explore more robust active belief dynamics that can better handle uncertainty and noise.

(2) Limited Exploration of Model Variants: Due to computational resource constraints, our experiments on prompting methods were primarily conducted using DeepSeek V3.2, and we did not extensively evaluate the approach across a broader range of LLM backbones. Furthermore, while our work addresses the fundamental challenge of belief updating and is expected to be compatible with various methods, we have not yet explored alternative designs to further facilitate accurate belief generation, such as integrating dense reward shaping or auxiliary supervision signals. Future work could incorporate these advanced designs to refine the belief intervention process.

\section*{Ethics Statement}

This work aims to develop LLM-based embodied agents within simulated environments. The VirtualHome and ALFWorld environment setup and related data strictly follow the specifications of VirtualHome \citep{puig2018virtualhome}, ALFWorld \citep{shridhar2020alfworld}, and ScienceWorld~\citep{wang2022scienceworld}. We utilize VirtualHome v2.3.0\footnote{\url{https://github.com/xavierpuigf/virtualhome/tree/master}} (MIT license\footnote{\url{https://github.com/xavierpuigf/virtualhome/blob/master/LICENSE}}), ALFWorld\footnote{\url{https://github.com/alfworld/alfworld}} (MIT license\footnote{\url{https://github.com/alfworld/alfworld/blob/master/LICENSE}}) and ScienceWorld\footnote{\url{https://github.com/allenai/ScienceWorld}} (MIT license\footnote{\url{https://github.com/allenai/ScienceWorld/blob/main/LICENSE}}) to conduct our experiments. 
All the LLMs we use for fine-tuning are open-source, and we strictly follow the protocols for the academic use of these models.
Additionally, while AI assistants (e.g., Cursor and ChatGPT) were partially utilized for code optimization and linguistic refinement, we affirm that all core content and findings in this paper are the original work of the authors.

\section*{Acknowledgements} 
This work was supported by the Research Grants Council of Hong Kong (15209724, 15205325), and also in part by the PolyU Postdoc Matching Fund Scheme (4-W40Z). The authors would like to thank the anonymous reviewers for their valuable feedback and constructive suggestions.

\bibliography{custom}

\appendix

\newpage

\section{Reliability Evaluation}
\label{app:probing_reliability}

To verify that our probing method accurately reflects the agent's true internal belief state—rather than generating hallucinations or unrelated outputs—we conducted a consistency analysis. We hypothesize that if the probing method is reliable, the belief elicited by the probe at timestep $t$ should be highly consistent with the agent's explicit reasoning (Thought) manifested in the subsequent timestep.

We validate the reliability by comparing the probing results against the agent's subsequent behavior. The specific procedure is as follows:

\begin{itemize}
    \item \textbf{Probing at $t$:} At a given timestep $t$, we focus on a specific object variable $o_t$. We query the agent's belief upon receiving obervation $o_t$ using our probing method to obtain a binary answer (Yes or No), denoted as the \textit{Predicted Label}.
    \item \textbf{Action at t:} We then allow the agent to process observation $o_t$ and conduct next step reasoning, denoted as $a_{t}$. We extract the agent's explicit understanding regarding the specific object variable $o_t$ from the thought component of $a_{t}$, denoted as the \textit{True Label}.
    \item \textbf{Comparison:} We compare the consistency between the probed answer and the subsequent thought across 100 sampled cases.
\end{itemize}

\begin{figure}[ht]
    \centering
    \includegraphics[width=0.85\linewidth]{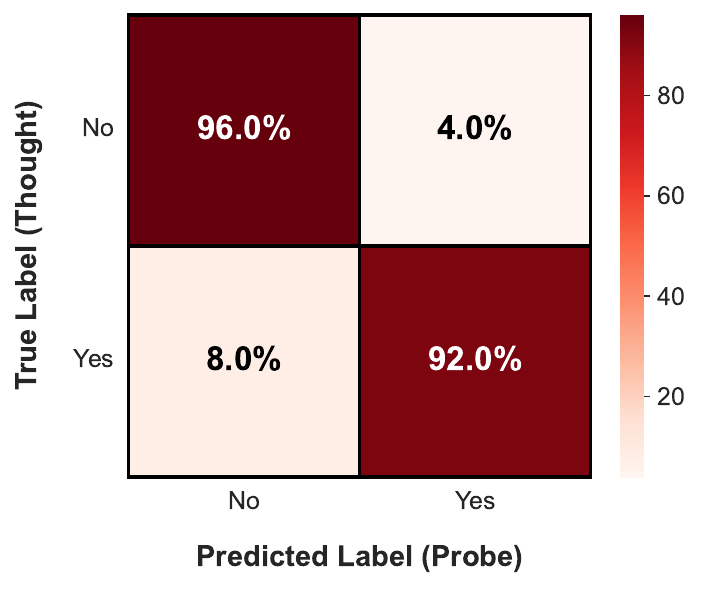}
    \caption{Confusion Matrix evaluating the consistency between the Probed Answer (Predicted Label) and the Agent's Subsequent Thought (True Label). The high values on the diagonal indicate that the probing method reliably reflects the agent's internal beliefs.}
    \label{fig:probing_reliability}
\end{figure}

We visualized the consistency between the probed beliefs and the agent's thoughts using a confusion matrix.
As illustrated in Figure~\ref{fig:probing_reliability}, our probing method demonstrates a high degree of fidelity with the agent's internal reasoning at next step:
\begin{itemize}
    \item When the agent's subsequent thought indicates a negative state (\textit{No}), the probe correctly identifies this belief \textbf{96.0\%} of the time.
    \item When the agent's subsequent thought indicates a positive state (\textit{Yes}), the probe correctly aligns with this belief \textbf{92.0\%} of the time.
\end{itemize}

The low off-diagonal error rates (4.0\% and 8.0\%) indicate minimal discrepancy. This strong alignment confirms that our method is reliable: it successfully externalizes the agent's latent beliefs without significant distortion, validating its utility for interpreting the agent's decision-making process.

\section{Datasets and Preprocessing}
\label{app:datasets}


\textbf{ALFWorld} is an interactive text-based environment that parallels the embodied worlds found in the ALFRED dataset. In this domain, agents are tasked with exploring a simulated household to complete high-level instructions, such as put a clean apple in the fridge.'' The dataset includes both seen'' splits for in-distribution evaluation and ``unseen'' splits to test out-of-distribution generalization. For the ReAct baseline, we utilize the SFT data generated by \citet{song2024trial}.

\textbf{ScienceWorld} is a complex text-based virtual environment designed to simulate elementary science experiments. It encompasses various distinct task types, such as thermodynamics and electrical circuits, requiring agents to ground their understanding of scientific concepts through practical, embodied interaction. Similar to ALFWorld, we adopt the training trajectories for the ReAct baseline provided by the \citet{song2024trial}. To ensure computational efficiency, we Task-8, Task-9, Task-1 (boil/freeze), Task-4 (grow fruit/plant), Task-5 (clean), Task-7 (paint), and Task-10 (decorate) due to their excessively long task-solving trajectories.



\textbf{VirtualHome} is a platform that simulates complex daily household activities, where agents execute programs to interact with objects and the environment. Unlike the previous datasets, we align the SFT data for the ReAct baseline and the experimental setup with the methodology established in STeCa~\citep{wang2025steca}. Consistent with the STeCA setting, we filter the dataset significantly; specifically, we remove approximately half of the training and testing data instances. This reduction is performed because many tasks in the original dataset are highly similar, ensuring a more efficient evaluation.

Across all three domains, the environments provide binary final rewards, where a reward of 1 indicates successful task completion and 0 indicates failure. Consequently, we report the average reward across the tested tasks as the success rate.

\textbf{Dataset Statistics} We summarize the detailed statistics of the three datasets in Table~\ref{tab:dataset_statistics}. The table reports the number of instances in the training set, as well as the Test-Seen'' (in-distribution) and Test-Unseen'' (out-of-distribution) evaluation sets. It also lists the average number of interaction turns required for expert trajectories, which serves as an indicator of task complexity across the different environments.

\begin{table}[ht]
    \centering
    \resizebox{\columnwidth}{!}{%
        \begin{tabular}{lcccc}
            \toprule
            \textbf{Dataset} & \textbf{\#Train} & \textbf{\#Test-Seen} & \textbf{\#Test-Unseen} & \textbf{\#Turns} \\
            \midrule
            ALFWorld     & 2851 & 140 & 134 & 7.97 \\
            VirtualHome  & 2460 & 125 & 125 & 8.79 \\
            ScienceWorld & 1253 & 151 & 161 & 9.64 \\
            \bottomrule
        \end{tabular}%
    }
    \caption{Statistics of datasets. ``Test-Seen'' and ``Test-Unseen'' are test set with seen and unseen scenarios respectively. ``\#Turns'' denotes the average number of interaction turns for the expert trajectories.}
    \label{tab:dataset_statistics}
\end{table}

\section{Additional Details about Baselines}
\label{app:baseline}

In this section, we provide additional implementation details for the baseline methods, categorized by their prompting strategies and training methodologies.

\paragraph{Prompting Settings.} We consider three frameworks: (1) No-Thinking\citep{ma2025reasoning}: The agent generates an action directly at each time step without any reasoning step. (2) Plan-and-Act\citep{kim2025reflact}: The agent conducts reasoning only at the first step and outputs actions without further thoughts in subsequent steps. and (3) ReAct~\citep{yao2022react}: The agent first reasons about the next action at each time step and then generates an action.

\paragraph{Training Settings.} We employ three distinct approaches: (1) SFT\citep{chen2023fireact}: The model is fine-tuned using standard supervised learning on a dataset of expert trajectories. (2) PPO\citep{schulman2017proximal}: A proximal policy optimization algorithm that utilizes a separate value network (critic) to reduce variance and stabilize training. and (3) GRPO~\citep{shao2024deepseekmath}: A group relative policy optimization method that eliminates the need for a critic model by estimating the baseline from the average reward of a group of sampled outputs for the same input.

\section{Additional Comparisons with Advanced Baselines}
\label{app:advanced_baselines}

we provide additional comparisons with more advanced baselines to further clarify the role of EVU. Specifically, we aim to answer two questions: (1) whether the gain of EVU mainly comes from better context management or from its active belief intervention mechanism, and (2) whether EVU can also benefit stronger agentic frameworks beyond a standard ReAct-style policy. To this end, we conduct three additional experiments: a short-context control experiment, a comparison against a history-summarization baseline, and an integration with a search-based agentic planner.

\textbf{Is EVU more than context management?}
To disentangle belief inertia from long-context crowding, we first revisit the failure cases identified in our analysis and truncate the interaction history to retain only the most recent two turns, so that the critical observation remains explicitly visible and no long-context retrieval is required. Even under this short-context setting, 95\% of the cases still exhibit belief inertia, where the agent ignores the immediate contradictory observation and continues to follow its prior belief. This result suggests that belief inertia is not merely a symptom of crowded context, but a distinct failure to integrate contradictory evidence. We further compare EVU against MEM1~\cite{zhou2025mem1}, a representative history-summarization baseline that manages context through memory compression but does not include an explicit estimate-verify-update loop. As shown in Table~\ref{tab:mem1_compare}, EVU consistently outperforms MEM1 on ScienceWorld, improving performance from 67.1 to 70.9 on seen tasks and from 64.9 to 70.8 on unseen tasks. These results indicate that passive context management alone is insufficient to overcome belief inertia; the key benefit comes from actively forcing the agent to estimate, verify, and update its belief state.

\begin{table}[ht]
\centering
\small
\begin{tabular}{lcc}
\toprule
Method & Seen & Unseen \\
\midrule
MEM1 & 67.1 & 64.9 \\
EVU (Ours) & 70.9 & 70.8 \\
\bottomrule
\end{tabular}
\caption{Comparison with the history-summarization baseline MEM1 on ScienceWorld. EVU consistently outperforms passive context management on both seen and unseen tasks.}
\label{tab:mem1_compare}
\end{table}

\textbf{Can EVU benefit stronger agentic planners?}
To evaluate whether EVU is compatible with more advanced agentic systems, we integrate EVU with LLM-MCTS~\cite{zhao2023large} on VirtualHome. This experiment examines whether EVU is complementary to search-based planning rather than being tied to a simple action-generation policy. As shown in Table~\ref{tab:mcts_compare}, incorporating EVU consistently improves over the strong LLM-MCTS baseline, raising performance from 28.8 to 31.2 on seen tasks and from 25.6 to 28.8 on unseen tasks. This result shows that EVU acts as a generalizable cognitive module that complements advanced planning algorithms by helping the agent maintain a more accurate belief state during search and execution, rather than being useful only in a standalone intervention setting.

\begin{table}[ht]
\centering
\small
\begin{tabular}{lcc}
\toprule
Method & Seen & Unseen \\
\midrule
LLM-MCTS & 28.8 & 25.6 \\
LLM-MCTS + EVU (Ours) & 31.2 & 28.8 \\
\bottomrule
\end{tabular}
\caption{Comparison with the search-based agentic planner LLM-MCTS on VirtualHome. EVU consistently improves over a stronger planning baseline.}
\label{tab:mcts_compare}
\end{table}

\section{Additional Implementation Details}
\label{app:more_implementation}

\begin{table}[t]
    \centering
    \small
    \begin{tabular}{lc}
        \toprule
        \textbf{Hyperparameter} & \textbf{Value} \\
        \midrule
        \textit{SFT Phase} & \\
        \quad Learning Rate & $1\mathrm{e}{-5}$ \\
        \quad Scheduler & Cosine \\
        \quad Epochs & 3 \\
        \midrule
        \textit{RL Phase} & \\
        \quad Group Size ($G$) & 6 \\
        \quad Learning Rate & $1 \times 10^{-6}$ \\
        \quad Total Steps & 250 \\
        \quad KL Coefficient ($\beta$) & 0.01 \\
        \quad Temperature (rollout) & 0.8 \\
        \quad Temperature (validation) & 0.0 \\
        \midrule
        \textit{Common / Other} & \\
        \quad Max Prompt Length & 5120 \\
        \quad Max Response Length & 512 \\
        \quad Micro Batch Size & 4 \\
        \quad Mini Batch Size & 32 \\
        \quad Invalid Action Penalty & 0.1 \\
        \quad Gradient Checkpointing & True \\
        \quad Tensor Parallel Size & 2 \\
        \quad Temperature (eval) & 0.0 \\
        \bottomrule
    \end{tabular}
    \caption{Key hyperparameters for SFT and RL training.}
    \label{tab:grpo_params}
\end{table}

Our training infrastructure for both Supervised Fine-Tuning (SFT) and Reinforcement Learning (RL) is built upon the verl-agent framework~\citep{feng2025group}.
All key hyperparameters for both phases are summarized in Table~\ref{tab:grpo_params}.
The templates corresponding to the ALFWorld, ScienceWorld, and VirtualHome environments are illustrated in Figure \ref{fig:template_alfworld}, Figure \ref{fig:template_vh}, and Figure \ref{fig:template_sciworld}, respectively.

\section{Experimental Setup for Different Belief Intervention Strategies}
\label{app:variant_details}

In this section, we provide a detailed formulation of the belief intervention strategies employed in our experiments. We first define a general framework for belief intervention and then describe how each specific strategy instantiates this framework.

\subsection{General Formulation}

We consider an agent interacting with an environment to achieve a goal $G$. At any time step $t$, given the interaction history and current observation (collectively denoted as Context $C_t$), the standard agent directly generates a thought chain and an action. 

To investigate the role of environmental reasoning, we introduce an intermediate \textbf{Belief Intervention (BI)} module. The decision-making process is decomposed into two phases:
\begin{enumerate}
    \item \textbf{Belief Generation:} The agent first generates a belief state content $\mathcal{B}$ based on a specific intervention strategy $\mathcal{S}$.
    \item \textbf{Action Generation:} The agent then generates the thought and action $A_t$ conditioned on both the context and the generated belief.
\end{enumerate}

Formally, this process can be represented as:
\begin{equation}
    C_t \xrightarrow{\text{Strategy } \mathcal{S}} \mathcal{B}_t \xrightarrow{} A_t
\end{equation}
where $\mathcal{B}_t$ varies depending on the definition of the strategy $\mathcal{S}$.

\subsection{Strategy Instantiations}

Below, we detail the five strategies compared in the main text. To illustrate the differences, we provide a running example where the agent's goal is to \textit{"Put the apple in the fridge"}, and the agent is currently standing in front of a \textbf{closed fridge} holding an apple.

\paragraph{NoneBI (Baseline).}
This represents the standard setting without any explicit belief intervention. The agent proceeds directly from observation to action.
\begin{itemize}
    \item \textbf{Formulation:} $\mathcal{B}_t = \emptyset$ (Empty set).
    \item \textbf{Mechanism:} The thought-action generation is solely dependent on the context $C_t$.
\end{itemize}

\begin{quote}
\textbf{Example Output:} \\
\textit{Thought:} The fridge is closed. I need to open it to put the apple inside. \\
\textbf{Action:} \texttt{Open(Fridge)}
\end{quote}

\paragraph{DBI (Direct Belief Intervention).}
DBI requires the agent to explicitly generate the current belief state regarding the environment before planning.
\begin{itemize}
    \item \textbf{Formulation:} $\mathcal{B}_t = S_t$, where $S_t$ is the description of the current environmental state relevant to the goal.
\end{itemize}

\begin{quote}
\textbf{Example Output:} \\
\textbf{[Belief]:} The fridge is currently closed. I am holding the apple. \\
\textit{Thought:} Since the fridge is closed, I must open it first. \\
\textbf{Action:} \texttt{Open(Fridge)}
\end{quote}

\paragraph{OBI (Observation-based Belief Intervention).}
OBI compels the agent to ground its belief in raw perception. It forces the agent to reiterate the recent observation text explicitly before summarizing the belief state.
\begin{itemize}
    \item \textbf{Formulation:} $\mathcal{B}_t = \{O_t \oplus S_t\}$, where $O_t$ is the reiteration of the observation and $S_t$ is the inferred state.
\end{itemize}

\begin{quote}
\textbf{Example Output:} \\
\textbf{[Belief]:} \textit{Observation:} I see a white fridge door in front of me. \textit{State:} The fridge is closed and I cannot see inside. \\
\textit{Thought:} I need to access the interior of the fridge. \\
\textbf{Action:} \texttt{Open(Fridge)}
\end{quote}

\paragraph{ReflAct.}
ReflAct~\citep{kim2025reflact} encourages the agent to reflect on its dynamic progress rather than just the static state. It prompts the agent to evaluate what has been accomplished and what remains to be done relative to the goal $G$.
\begin{itemize}
    \item \textbf{Formulation:} $\mathcal{B}_t = \text{Reflect}(S_t, G)$, focusing on the gap between current state and goal.
\end{itemize}

\begin{quote}
\textbf{Example Output:} \\
\textbf{[Belief]:} I have successfully picked up the apple. However, the goal is to put it in the fridge, and the fridge is currently closed. \\
\textit{Thought:} The next logical step is to open the fridge. \\
\textbf{Action:} \texttt{Open(Fridge)}
\end{quote}

\paragraph{VAGEN.}
VAGEN~\citep{wang2025vagen} tasks the agent with a predictive "look-ahead" process. The agent must identify the current belief state, propose a hypothetical action, and then predict the resulting environmental state.
\begin{itemize}
    \item \textbf{Formulation:} $\mathcal{B}_t = \{S_t \xrightarrow{\text{Hypothetical Action } \tilde{a}} S_{t+1}\}$, representing a mental simulation of the immediate future.
\end{itemize}

\begin{quote}
\textbf{Example Output:} \\
\textbf{[Belief]:} \textit{Current State:} The fridge is closed. \textit{Hypothetical Action:} Open the fridge. \textit{Predicted Future State:} The fridge door will be open, revealing the contents inside. \\
\textit{Thought:} This action will allow me to place the apple inside. \\
\textbf{Action:} \texttt{Open(Fridge)}
\end{quote}


\section{Experimental Setup for Belief Inertia Mitigation Analysis}
\label{app:belief_intertia_mitigation}

To investigate the internal mechanism of \model in mitigating belief inertia, we conduct a fine-grained probing analysis using the identical set of \textit{observational neglect} cases described in Section 3.2. These cases represent critical moments where the agent's action $a_t$ fails to reflect the current observation $o_t$, indicating a strong inertia from prior beliefs. 

We track the belief evolution by probing the agent at four distinct stages of our active belief dynamic process. For each stage, we construct the input prompt by progressively accumulating the intermediate reasoning outputs. Specifically, the probing stages are defined as follows: (1) \textbf{Current Belief}: This represents the baseline state prior to any active intervention, probed using the historical context $(B_{t-1}, a_{t-1}, o_{t-1})$; (2) \textbf{Estimated Belief}: This captures the belief state immediately after the estimation phase, where the context is augmented with the generated estimation to form $(B_{t-1}, a_{t-1}, o_{t-1}, E_t)$; (3) \textbf{Verification Belief}: This reflects the state after the agent validates the estimation against the observation, probed under the context $(B_{t-1}, a_{t-1}, o_{t-1}, E_t, V_t)$; and (4) \textbf{Updated Belief}: This represents the final consolidated state after the update phase, probed using the complete context $(B_{t-1}, a_{t-1}, o_{t-1}, E_t, V_t, B_t)$. By comparing the belief values across these stages, we quantify the contribution of each component in correcting the belief inertia.


\section{Experimental Setup for Task Difficulty Analysis}
\label{app:task_difficulty}

To systematically evaluate the robustness of active belief dynamics as tasks become more challenging, we conduct a difficulty analysis within the ALFWorld environment. We classify task difficulty based on the minimum number of subgoals required to achieve the final objective, as tasks with more subgoals necessitate longer interaction horizons. We obtain four distinct difficulty levels: \textit{Easy} (0--4 subgoals), \textit{Medium} (5--8 subgoals), \textit{Hard} (9--12 subgoals), and \textit{Very Hard} (13--16 subgoals). 
We randomly sampled a total of 200 tasks to ensure diverse coverage across these difficulty levels. For each level, we compare the success rate of our proposed method against three training baselines: SFT, GRPO, and PPO. 
To quantify the advantage, we calculate the relative success rate improvement ($\Delta$ Success Rate) of our method over each baseline.

\begin{figure*}[t] 
    \centering
    \begin{tcolorbox}[
        title=\textbf{ALFWorld Prompt Template}, 
        colframe=blue!50!black,  
        colback=blue!5!white,    
        coltitle=white,          
        fonttitle=\bfseries,     
        arc=1mm,                 
        boxsep=2pt,              
        fontupper=\small         
    ]
        Interact with a household to solve a task. Imagine you are an intelligent agent in a household environment and your target is to perform actions to complete the task goal. 

        At each step, you will be given task goal, action history and the last turn's information (Reason, Belief State, Thought, and Action).

        \textbf{You need to process the information in a specific order:}
        \begin{enumerate}[nosep, leftmargin=*] 
            \item \textbf{Reason}: Analyze the last action and the observation in one or two concise sentences. What did you expect to see? What did you actually see? Does this confirm or contradict your previous belief?
            \item \textbf{Belief State}: State where the agent is, what it is holding, and the known status of goal-related objects. Do NOT list irrelevant objects.
            \item \textbf{Thought}: Plan your future actions based on the updated belief.
            \item \textbf{Action}: Output your next action.
        \end{enumerate}
        
        \vspace{4pt} 
        \textbf{The available actions are:}
        \begin{multicols}{3} 
            \begin{enumerate}[nosep, leftmargin=*, label=\arabic*.]
                \item go to (recep)
                \item task (obj) from (recep)
                \item put (obj) in/on (recep)
                \item open (recep)
                \item close (recep)
                \item toggle (obj) (recep)
                \item clean (obj) with (recep)
                \item heat (obj) with (recep)
                \item cool (obj) with (recep)
            \end{enumerate}
        \end{multicols}
        \textit{where (obj) and (recep) correspond to objects and receptacles.}
        
        After your each turn, the environment will give you immediate feedback based on which you plan your next few steps. If the environment output ``Nothing happened'', that means the previous action is invalid and you should try more options.

        \vspace{4pt}
        \textbf{Your response should use the following format:}
        \begin{tcolorbox}[colback=white, colframe=gray!30, arc=0mm, boxsep=0pt, left=4pt, top=2pt, bottom=2pt]
            \ttfamily 
            Reason: <Analyze expectation vs. actual observation to update your understanding>\\
            Belief State: <your belief state>\\
            Thought: <your thoughts>\\
            Action: <your next action>
        \end{tcolorbox}

        Your task is to complete the task goal: \{task\_goal\}

        Below is the action history and the last turn's information:

        \textbf{Action History:} \{action\_history\}

        \textbf{Last Turn's Information:} \{last\_turn\_information\}
        
    \end{tcolorbox}
    \caption{Prompt template of our method on the ALFWorld benchmark.}
    \label{fig:template_alfworld}
\end{figure*}

\begin{figure*}[t]
    \centering
    \begin{tcolorbox}[
        title=\textbf{VirtualHome Prompt Template}, 
        colframe=blue!50!black,  
        colback=blue!5!white,    
        coltitle=white,          
        fonttitle=\bfseries,     
        arc=1mm,                 
        boxsep=2pt,              
        fontupper=\small         
    ]
        Interact with a household to solve a task. Imagine you are an intelligent agent in a household environment and your target is to perform actions to complete the task goal. At the beginning of your interactions, you will be given the detailed description of the current environment and your goal to accomplish.

        At each step, you will be given task goal, action history and the last turn's information (Reason, Belief State, Thought, and Action).

        \textbf{You need to process the information in a specific order:}
        \begin{enumerate}[nosep, leftmargin=*]
            \item \textbf{Reason}: Analyze the last action and the observation in one or two concise sentences. What did you expect to see? What did you actually see? Does this confirm or contradict your previous belief?
            \item \textbf{Belief State}: State where the agent is, what it is holding, and the known status of goal-related objects. Do NOT list irrelevant objects.
            \item \textbf{Thought}: Plan your future actions based on the updated belief.
            \item \textbf{Action}: Output your next action.
        \end{enumerate}
        
        \vspace{4pt}
        \textbf{The available actions are:}
        \begin{multicols}{3} 
            \begin{enumerate}[nosep, leftmargin=*, label=\arabic*.]
                \item walk to (obj)
                \item run to (obj)
                \item grab (obj)
                \item open (obj)
                \item close (obj)
                \item put (obj) on (recep)
                \item put (obj) in (recep)
                \item switch on (obj)
                \item switch off (obj)
                \item drink (obj)
                \item look at (obj)
                \item sit on (obj)
                \item stand up
                \item watch (obj)
                \item wipe (obj)
                \item type on (obj)
                \item wash (obj)
                \item cut (obj)
                \item eat (obj)
                \item sleep
                \item wake up
                \item plug in (obj)
                \item plug out (obj)
                \item pour (obj) into (recep)
                \item move (obj)
                \item release
                \item turn to (obj)
            \end{enumerate}
        \end{multicols}
        
        After your each turn, the environment will give you immediate feedback based on which you plan your next few steps. If the environment output ``Nothing happened'', that means the previous action is invalid and you should try more options.

        \vspace{4pt}
        \textbf{Your response should use the following format:}
        \begin{tcolorbox}[colback=white, colframe=gray!30, arc=0mm, boxsep=0pt, left=4pt, top=2pt, bottom=2pt]
            \ttfamily
            Reason: <Analyze expectation vs. actual observation to update your understanding>\\
            Belief State: <your belief state>\\
            Thought: <your thoughts>\\
            Action: <your next action>
        \end{tcolorbox}

        Your task is to complete the task goal: \{task\_goal\}

        Below is the action history and the last turn's information:

        \textbf{Action History:} \{action\_history\}

        \textbf{Last Turn's Information:} \{last\_turn\_information\}
        
    \end{tcolorbox}
    \caption{Prompt template of our method on the VirtualHome benchmark.}
    \label{fig:template_vh}
\end{figure*}

\begin{figure*}[t]
    \centering
    \begin{tcolorbox}[
        title=\textbf{ScienceWorld Prompt Template}, 
        colframe=blue!50!black,  
        colback=blue!5!white,    
        coltitle=white,          
        fonttitle=\bfseries,     
        arc=1mm,                 
        boxsep=2pt,              
        fontupper=\small         
    ]
        You are a helpful assistant to do some scientific experiment in an environment.
        In the environment, there are several rooms: kitchen, foundry, workshop, bathroom, outside, living room, bedroom, greenhouse, art studio, hallway.
        You should explore the environment and find the items you need to complete the experiment.
        You can teleport to any room in one step.
        All containers in the environment have already been opened, you can directly get items from the containers.

        At each step, you will be given task goal, action history and the last turn's information (Reason, Belief State, Thought, and Action).

        \textbf{You need to process the information in a specific order:}
        \begin{enumerate}[nosep, leftmargin=*]
            \item \textbf{Reason}: Analyze the last action and the observation in one or two concise sentences. What did you expect to see? What did you actually see? Does this confirm or contradict your previous belief?
            \item \textbf{Belief State}: State where the agent is, what it is holding, and the known status of goal-related objects. Do NOT list irrelevant objects.
            \item \textbf{Thought}: Plan your future actions based on the updated belief.
            \item \textbf{Action}: Output your next action.
        \end{enumerate}
        
        \vspace{4pt}
        \textbf{The available actions are:}
        \begin{multicols}{2} 
            \begin{itemize}[nosep, leftmargin=*, label=\textbullet]
                \item \texttt{open OBJ}: open a container
                \item \texttt{close OBJ}: close a container
                \item \texttt{activate OBJ}: activate a device
                \item \texttt{deactivate OBJ}: deactivate a device
                \item \texttt{connect OBJ to OBJ}: connect electrical components
                \item \texttt{disconnect OBJ}: disconnect electrical components
                \item \texttt{use OBJ [on OBJ]}: use a device/item
                \item \texttt{look around}: describe the current room
                \item \texttt{examine OBJ}: describe an object in detail
                \item \texttt{look at OBJ}: describe a container's contents
                \item \texttt{read OBJ}: read a note or book
                \item \texttt{move OBJ to OBJ}: move an object to a container
                \item \texttt{pick up OBJ}: move an object to the inventory
                \item \texttt{pour OBJ into OBJ}: pour a liquid into a container
                \item \texttt{mix OBJ}: chemically mix a container
                \item \texttt{teleport to LOC}: teleport to a specific room
                \item \texttt{focus on OBJ}: signal intent on a task object
                \item \texttt{wait}: task no action for 10 steps
                \item \texttt{wait1}: task no action for a step
            \end{itemize}
        \end{multicols}

        \vspace{4pt}
        \textbf{Your response should use the following format:}
        \begin{tcolorbox}[colback=white, colframe=gray!30, arc=0mm, boxsep=0pt, left=4pt, top=2pt, bottom=2pt]
            \ttfamily
            Reason: <Analyze expectation vs. actual observation to update your understanding>\\
            Belief State: <your belief state>\\
            Thought: <your thoughts>\\
            Action: <your next action>
        \end{tcolorbox}

        Your task is to complete the task goal: \{task\_goal\}

        Below is the action history and the last turn's information:

        \textbf{Action History:} \{action\_history\}

        \textbf{Last Turn's Information:} \{last\_turn\_information\}
        
    \end{tcolorbox}
    \caption{Prompt template of our method on the ScienceWorld benchmark.}
    \label{fig:template_sciworld}
\end{figure*}

\end{document}